# Assigning Species Information to Corresponding Genes by a Sequence Labeling Framework

Ling Luo[+], Chih-Hsuan Wei[+], Po-Ting Lai, Qingyu Chen, Rezarta Islamaj Doğan, Zhiyong Lu[*]

National Center for Biotechnology Information (NCBI), National Library of Medicine (NLM), National Institutes of Health (NIH), Bethesda, MD 20894, USA
*Corresponding author: zhiyong.lu@nih.gov

## ABSTRACT

The automatic assignment of species information to the corresponding genes in a research article is a critically important step in the gene normalization task, whereby a gene mention is normalized and linked to a database record or identifier by a text-mining algorithm. Existing methods typically rely on heuristic rules based on gene and species co-occurrence in the article, but their accuracy is suboptimal. We therefore developed a high-performance method, using a novel deep learning-based framework, to classify whether there is a relation between a gene and a species. Instead of the traditional binary classification framework in which all possible pairs of genes and species in the same article are evaluated, we treat the problem as a sequence-labeling task such that only a fraction of the pairs needs to be considered. Our benchmarking results show that our approach obtains significantly higher performance compared to that of the rule-based baseline method for the species assignment task (from 65.8% to 81.3% in accuracy). The source code and data for species assignment are freely available at https://github.com/ncbi/SpeciesAssignment.

## 1   INTRODUCTION

With the rapid growth of biomedical literature, automatically extracting and summarizing the knowledge in the literature becomes increasingly important to biomedical research in the areas such as biocuration assistance [1, 2] and COVID research [3]. The gene is one of the most important key concepts in biomedical research and is relevant to genetic variation, pharmacogenomics, cancer research, and precision medicine. Many text-mining studies [4-6] rely highly on the automatic extraction of gene names in the text. Because multiple genes may share the same name, mapping gene names to unique concept identifiers is a very important step. NCBI Gene is a major database for gene records. Many studies [7-9] focus on mapping the gene names to the gene identifiers, and this task is widely known as *gene name normalization* (or *gene linking*).

---

[+] The authors wish it to be known that, in their opinion, the first two authors should be regarded as joint first authors.

The ortholog gene in different species, however, is associated with different gene identifiers, which exacerbates the difficulty of the gene normalization task. One means to decrease ambiguity is to identify the corresponding species of each gene mention (termed *species assignment*) and aim to narrow down the candidates for the possible gene identifiers. Few existing tools [10-12], however, have sufficient accuracy in regard to the species name recognition task, and none was designed to disambiguate the genes to the corresponding species. Further, few studies have developed rule-based methods [13-15] to disambiguate the corresponding species of the gene, based on co-occurrence in the same sentence or paragraph. The most popular strategies to determine the corresponding species include the use of: (1) the most nearby species of the gene in the same sentence; (2) the most frequent species in the same paragraph; (3) the corresponding species of the gene prefix that is represented (e.g., "**h**CB1R" to human); and (4) the species in the title. One of our previous works, SR4GN [15], was designed to recognize the corresponding species by leveraging the rules noted above to obtain good performance. SR4GN was successfully embedded in a precise gene tagger, GNormPlus [7], and applied to the entire PubMed and PMC [6] databases for gene recognition and normalization. SR4GN, however, frequently failed to find the correct corresponding species of the gene mentions when no species was mentioned within the same sentence. In addition, SR4GN missed >90% species of the entire NCBI taxonomy repository, as only a small portion of the species corresponds to the genes recorded in the NCBI gene database.

As we learned from previous studies, two main challenges remained. (1) Most of the existing species-recognition methods were not designed for gene normalization. In particular, some species-sensitive concepts (e.g., cell line, species strain) that are also helpful to species assignment and gene normalization were ignored. (2) Although the rule-based method used to assign the species to gene mentions is straightforward, it does not work when the corresponding species is not in proximity of the gene. This is especially the case when an article mentions multiple species, as it increases the difficulty of the task. For example, when a study uses multiple animal models to observe the expression of the human gene under defined criteria, this can cause difficulties even for manual assignment. To address the problem, we propose a novel species assignment approach based on deep learning via a sequence labeling framework. To the best of our knowledge, this work is the first that explores deep learning methods to assign species to gene mentions. The main contributions of our work can be summarized as follows:

- We develop a dictionary-based species tagger with state-of-the-art performance (94.3%, F-measure).

- We explore machine learning-based methods for the species assignment task. Instead of a traditional binary classification framework, we propose a novel species assignment approach based on a sequence labeling framework. We apply cutting-edge biomedical pre-trained language models (i.e., PubmedBERT [16] and Bioformer [17]) for both frameworks and improve performance from 65.8% to 81.3% in accuracy.

- We comprehensively compare the binary classification framework-based deep learning method, sequence labeling framework-based deep learning method, and existing rule-based method in the species assignment task.

## 2 METHODS

### 2.1 The workflow of the automatic species recognition and assignment

To address the two challenges, we proposed a new species tagger and a deep learning framework to optimize the performance of the species assignment. Figure 1 shows the architecture of our method, with end-to-end steps. The input data require free text with highlighted gene mentions. To precisely evaluate the performance of the species assignment but not the effect of another component, we provide the manually annotated gene mentions in the input. The two parts of our work include species recognition of species mentions, the species' concept identifier (NCBI taxonomy ID), and the species assignment, which links the corresponding species to gene mentions. Additional details are presented in the following section.

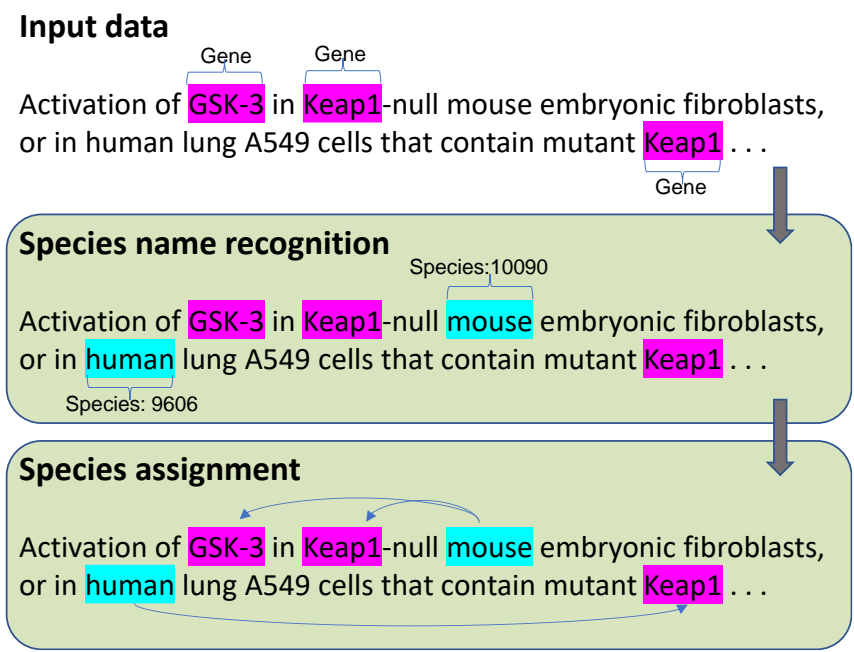

Figure 1: The architecture of the species recognition and assignment.

## 2.2 Species recognition

Machine learning-based named entity recognition methods always achieve much better performance than do dictionary-based methods on most bioconcepts (i.e., gene, disease, chemical). Nevertheless, performance of species recognition, using dictionary-based methods, is competitive with machine learning-based methods [18]. This is because the main challenge of species recognition is not term variation or ambiguity, as most of the species names in the text follow the nomenclature of the Carl Linnaeus standard system [19], and the species names in the text are standardized. Rather, the volume of the species taxonomy is critical. More than 2 million unique species (>16 million species names) are recorded in NCBI taxonomy[1] as of 2022. The number of the species names is larger than that of other popular bioconcepts (e.g., disease, chemicals). Such a supervised-learning method may not be able to maintain coverage of a large-scale data set without support from a species lexicon. In addition, although species names are required to be linked to concept identifiers (NCBI taxonomy IDs), none of the existing supervised learning-based species taggers can map the species names to the specific concept identifiers.

---

[1] https://www.ncbi.nlm.nih.gov/taxonomy

Thus, we generated a dictionary-based species tagger that can better handle the enormous size of the species lexicon, based on the hierarchical structure of the taxonomy system. More specifically, our species tagger was implemented by adopting a prefix tree to reorganize the species names within a highly efficient structure for a string search. In addition, the structure can easily maximize the capacity to recognize name variations and the abbreviations of strain prefixes (e.g., "str" and "substr" in "E. coli str. K-12 substr. MG1655"). In the prefix tree, every node is a token. The children of a node are the next words in the species name. Thus, the same token in different species names shares the same node. For example, "Escherichia" and "coli" are the shared prefix nodes of the "Escherichia coli str. K-12" and the "Escherichia coli str. BL21." The token "str." is a common abbreviation of the strain prefix, which can be skipped during the string search. "K-12" and "BL21" are two individual child nodes under the node "coli." As shown in the species name recognition module in Figure 2, the structure of the prefix tree-based dictionary perfectly presents a corresponding structure to the taxonomy hierarchical system.

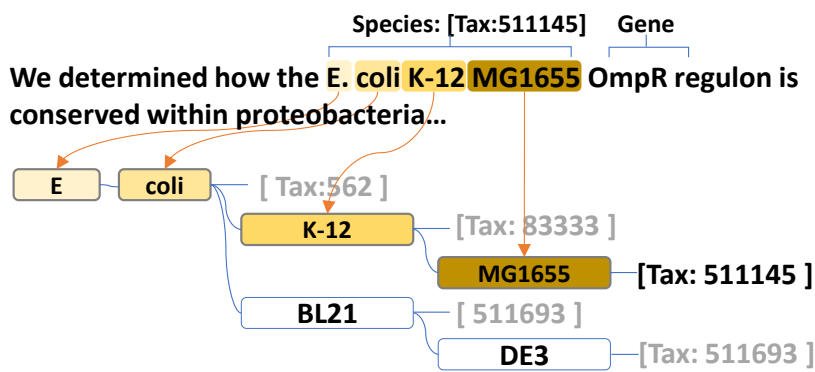

**Figure 2: Species name prefix tree and the name recognition in text.**

Due to the flexible design of the dictionary, our tagger also can solve several common term variations. (1) The strain prefixes (e.g., "str.", "substr.") can be simply recognized and skipped. Thus, there is no need to have a separate branch for the same species name with strain prefix. This not only saves space on the tree but also decreases search complexity. (2) The longer species name can be more easily recognized than that of another species with a shorter name. For example, "E. coli strain O157:H7" would be retrieved from the text, "Using a similar approach, we show that E. coli strain O157:H7 Stx prophage or prophage remnants invariably include paralogs of nanS often located downstream of the Shiga-like toxin genes" (PMID: 27481927), rather than "E. coli." In general, biomedical literature uses abbreviations to represent the concept name, with multiple tokens, but this is rare for species. Instead, genus names and strain codes are frequently used to represent parts of the species names. The genus name "Arabidopsis" represents "Arabidopsis thaliana" in PMID:31279220, and the species strain code (e.g., MG1655) can represent the specific strain of the species (e.g., E. coli str. k-12 substr. MG1655). To handle the case, we built a mapping from the genus names and strain codes to their corresponding species. Once a species (e.g., E. coli) is found in the text, the tagger searches the strain codes and the genus names via the species hierarchical system.

In addition, a frequent case was observed in our previous work [6]. Shortening the genus name by using the first capital letter (e.g., "Escherichia coli Escherichia to "E. coli") to represent the species occurs frequently. Sometimes, however, the genus is abbreviated by the first two letters (e.g., "Aedes aegypti" to "Ae. aegypti"). We expanded the species lexicon to cover such cases. We also recognize the cell line in the text, as it usually represents the animal model in vitro and in vivo.

## 2.3 Species assignment

The traditional rule-based method of species assignment relies highly on the species that are mentioned together in the surrounding context. The species, however, may not be in proximity to the gene. In addition, the case of multiple surrounding species is confusing in terms of the detection of the corresponding species of the gene name.

We deal with this task as a relation extraction between gene and species and establish a supervised machine learning-based method using biomedical transformer-based pre-trained language models (PLMs) (e.g., PubmedBERT [16] and Bioformer [17]) for this task. As an encoder to represent the input text, PLMs can measure the relevance between tokens (e.g., gene and species), which is then applied to various biomedical text mining tasks and can significantly surpass state-of-the-art performance. A straightforward framework that can recognize the corresponding species to the gene spans is the binary classification, which classifies each pair of gene and species. A positive outcome means that the gene corresponds to the species, and a negative one means that it does not. The binary classification method, however, is required to process all of the pairs between species and gene names, one by one, which is time-consuming, and it is difficult to handle large-scale data using advanced deep learning techniques. Moreover, these methods ignore the dependency between entity pairs, as it deconstructs the task into multiple independent entity pair classification subtasks.

Inspired by several previous works on relation extraction [20-22], we proposed a novel species assignment method based on the sequence labeling framework. As shown in Figure 3, we converted the task to a sequence labeling problem. Given an input text, the candidate entity (e.g., species entity of "mouse") in the text, the goal of the model is to recognize all corresponding genes (e.g., "GSK-3," "Keap1," and "phosphoinositide 3-kinase (PI3K)-protein kinase B)" at once. Therefore, the speed of the sequence labeling framework is significantly faster than the binary classification framework.

Two strategies to predict the corresponding species for gene mentions include (1) targeting the species to reach its belonging genes (S→G; species to gene), and (2) targeting the genes to reach the corresponding species (G→S; gene to species). S→G is much more efficient than is G→S, about 7 times faster, as the number of species is usually less than the gene mentions in the input text. In addition, S→G is slightly more accurate than is G→S. Next, we use the strategy of S→G to present the details of our sequence labeling framework. Specifically, to distinguish the gene and species from other tokens in the text, we inserted a pair of tags "<GENE>" and "</GENE>", in front of and at the end of the genes, and "<SPEC>" and "</SPEC>" in the same way for the species. In each iteration, we sequentially selected a species and assigned a pair of tags ("<ARG>" and "</ARG>") to distinguish the target species from the others. We further translated the tokens of the input text into a sequence within two statuses: "I" (inside), and "O" (outside), as the predicted sequence. The example in Figure 3 shows the architecture of our model. The genes that correspond to the target species (including the surrounding tags) are in "I" status, and other tokens and the genes that do not correspond to the target species are in "O" status. In the example, the input document contains two species (i.e., "mouse" and "human") and four gene mentions. The predicted sequence labels a gene to "I" status, indicating that the gene corresponds to the target species ("mouse"). At the end of the model, we used the softmax classification layer to summarize the probability scores of the labels of each token. We applied PubMedBERT [16] and Bioformer [17] as the PLMs models. PubMedBERT is the biomedical version of BERT and was recently created using only a biomedical vocabulary and datasets without transfer learning. Bioformer is a lightweight version of the traditional BERT model, which has been successfully applied in the biomedical domain. In most cases, each gene should be assigned to one species. Thus, we assigned the species with the highest predicted score to the genes, unless the two species with the highest score were in conjunction with each other (e.g., "*human and mouse cDNAs of ABCB9*"). In addition, we assigned the species with the highest scores to the genes when no corresponding species could be reached by the model.

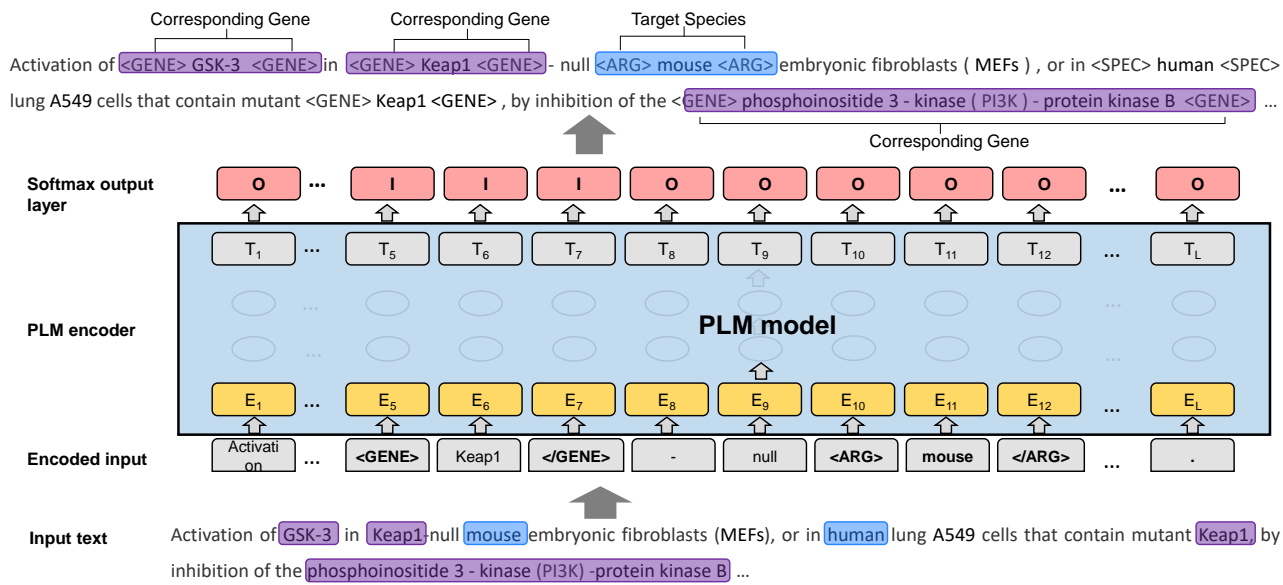

**Figure 3: The formulated labeled sequence and the PLM model.**

## 3 RESULTS

### 3.1 Evaluation of the species recognition

To better understand the performance of our species tagger, we first compared it with two other species taggers (i.e., Linnaeus [10] and SPECIES [12]) by their proposed corpora. Although many other species taggers [18, 23-25] obtained good performance with regard to the recognition of species boundary mentions, they did not address the normalization of the species. In fact, the normalization of species mentions for the species concepts (i.e., NCBI taxonomy IDs) is important to the species assignment task. Table 1 shows the performance of recognizing the species concept identifiers of individual taggers on Linnaeus and SPECIES corpora. The evaluation is at document-level, which means that, when one species appears multiple times, it should be counted only once. Our tagger attained the best performance as compared to the other two taggers on both of the corpora.

Table 1: Species normalization performance on Linnaeus and SPECIES corpora. The performances are in F-measures (the harmonic mean of the precision and recall).

| Corpus | # Articles | # Species | Our tagger | Linnaeus | SPECIES |
| --- | --- | --- | --- | --- | --- |
| Linnaeus | 100 full texts | 2851 | **94.3%** | 90.3% | 92.0% |
| SPECIES | 800 abstracts | 3708 | **82.7%** | 79.6% | 77.8% |

### 3.2 Evaluation of the species assignment

For the benchmarking of the species assignment, we chose two well-known gene-rich corpora (i.e., GNormPlus [7] and NLM-Gene [26]). Some articles, however, mentioned only one species (or no species). In such cases, all genes in the articles should link to the same species, without ambiguity (human is the species at default). To determine the extent of improvement with the new method, we need to focus on the articles' ambiguity issues. In that regard, the articles qualified for benchmarking require more than one species candidate for the gene mentions in the text. In addition, we excluded some articles that contained genes for which the corresponding species is in the full text but not in the abstract. Table 2 shows the number of abstracts in GNormPlus and NLM-Gene corpora that are qualified for benchmarking. Based on the criteria, fewer than half of the articles are eligible. The GNormPlus corpus is primarily focused on human genes, such that if a gene in an article corresponds to two or more species, only human genes are annotated in the corpus. To ensure that the evaluation can reflect the actual species diversity of the genes, we thus randomly selected the articles from the NLM-Gene corpus only for testing. The remaining eligible articles in the NLM-Gene and GNormPlus corpora were used for model development. In total, we collected 403 abstracts for training and 75 abstracts for evaluation.

Table 2: The corpora for benchmarking. The numbers in parentheses are the original numbers of the articles in individual corpus.

| Corpus | # Abstracts | Training | Test |
| --- | --- | --- | --- |
| GNrormPlus | 262 (694) | 262 | 0 |
| NLM-Gene | 216 (550) | 141 | 75 |
| Total | 478 (1245) | 403 | 75 |

In our experiments, we downloaded two biomedical PLMs (i.e., PubMedBERT[2], and Bioformer[3]) and evaluated them in both frameworks. The models were trained using the Adam [27] optimizer to minimize categorical cross-entropy loss. For parameter setting, we used PLMs with the default parameter settings and set the other hyper-parameters as follows: learning rate of 5e-6, batch size of 16. The number of training epochs was chosen by the early stopping strategy, according to the training loss score. We focused on evaluating the performance on species assignment, using the resampled corpus, and the manual curated gene mentions are given.

---

[2] https://huggingface.co/microsoft/BiomedNLP-PubMedBERT-base-uncased-abstract
[3] https://huggingface.co/bioformers/bioformer-cased-v1.0

To explore the effectiveness of our sequence labeling framework, we compared the performance of PLMs in classification and sequence labeling frameworks on the test. We also used the species-assignment module (a rule-based method) of GNormPlus as a baseline. Table 3 shows the results of different methods in terms of accuracy and processing time. As expected, deep learning methods provide better performance than does the rule-based method. Compared with the binary classification framework, our sequence labeling framework achieves similar or better performance. For processing time, our sequence labeling framework is more efficient. Specifically, it is 10–20 times faster than the binary classification framework when tested on GPUs and CPUs, respectively. Further, when comparing the two sequence-labeling strategies (S→G and G→S), we found that S→G is more efficient and accurate than is G→S for both PLM models (PubMedBERT and Bioformer). The highest performance (sequence labeling framework with the PubMedBERT model, using the S→G search strategy) increased about 16% compared to the baseline (GNormPlus), from 65.8% to 81.3%. Even though PubMedBERT achieved a slightly higher performance than did Bioformer, Bioformer is more efficient than is PubMedBERT in terms of both GPU and CPU environments (around two times faster) and may be a better option for processing the large-scale data sets (e.g., entire PubMed abstracts or PMC full texts).

**Table 3: The performance of the species assignment.**

| PLM Model | Framework | Strict-ACC | Relax-ACC | Processing time (Seconds per 100 abstracts) | |
|---|---|---|---|---|---|
| | | | | GPU | CPU |
| PubMedBERT | Sequence labeling (S→G) | **81.3%** | **85.4%** | 16 | 50 |
| Bioformer | | 80.7% | 83.7% | 10 | 25 |
| PubMedBERT | Sequence labeling (G→S) | 79.7% | 83.2% | 90 | 740 |
| Bioformer | | 78.1% | 82.5% | 60 | 320 |
| PubMedBERT | Binary classification | 79.8% | 83.6% | 140 | 1580 |
| Bioformer | | 78.1% | 82.7% | 80 | 674 |
| GNormPlus | Rule-based | 65.8% | 71.7% | 4 | 4 |

Note that all models were trained and evaluated on the same GPU (Tesla V100-SXM2-32GB) and CPU (Intel(R) Xeon(R) Gold 6226 CPU @ 2.70GHz, 24 Cores). Some genes may correspond to more than one species. Strict-ACC denotes strict accuracy, which requires that all the corresponding species of the gene should be extracted. Relax-ACC denotes relax accuracy, which accepts that only one corresponding species is extracted.

## 4 DISCUSSION

Despite our best efforts, there are still errors in the results of the species assignment. We examined all of the errors of the S→G sequence labeling framework using the Bioformer model (i.e., 83.7% accuracy) in the test set and grouped them into several major categories, as shown in Figure 4. In most cases, the nearest species have the highest probability to be the corresponding species of the gene spans. Thus, it is confusing to the machine if the surrounding species does not correspond to the gene. This situation causes 44% of our errors and is the major category of errors for our results. As the example in PMID:25277705, the species names of the respiratory syncytial virus (RSV) are glutted in the article, but the article concerns RSV infection, not the genetics of the virus. Thus, the corresponding species of the genes is human but not RSV. As we learned from this type of error, the genes of humans and viruses can be confused when there are two different research topics. In the first topic, the human gene function is relevant to the virus infection. In PMID:35238065, the human CLIC3 gene is a potential indicator of poor prognosis of hepatitis B virus-related acute-on-chronic liver failure. The other article, however, focuses on the variants of the virus sequence. In PMID:35416390, SARS-CoV-2 with E484K mutation in the spike gene is expressed in lower expected inhibitory activity of antibodies. To better address the issue, it is necessary to recognize the topic of the research.

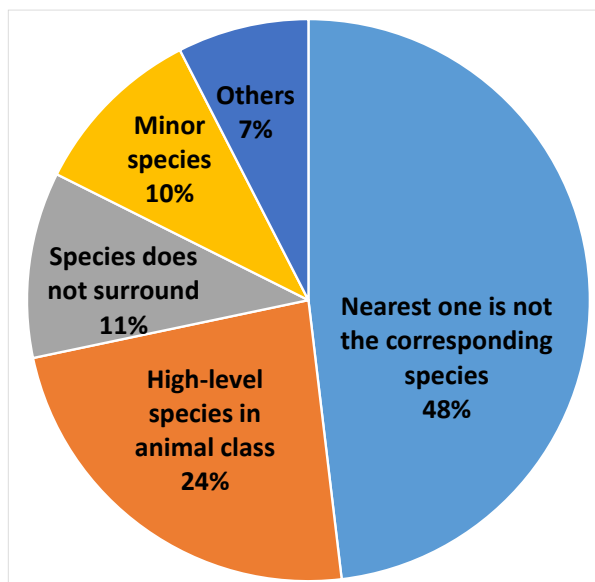

**Figure 4: The species assignment error types.**

The second error category is caused by a confusion of the species names within the higher level of the animal class. As an example, in PMID: 23195221, the "mammalian" represents human, mouse, and other mammals. The NLM-Gene corpus annotated the genes to mouse, however, as the full text analyzed the genes in the mouse model. Such an ortholog gene, which represents all of the belonging genes, exists widely in various literatures. No existing NCBI gene identifier, however, can represent those ortholog genes.

The third error type occurs when the corresponding species is far from the gene, and even the species is not in the same (or nearby) sentence(s). Sometimes, a closed species does not correspond to the gene. Such an example is seen in PMID20644716, the experiment in vitro/vivo that used a mouse model (cell line RAW 264.7) to understand the expression of the human ortholog gene. Unfortunately, our method cannot always handle the cases well. The other error type occurs when the number of the corresponding species in the article is much less than that of the other species. In such a case, our method may incorrectly assign the most frequently occurring species to the genes instead. For instance, the species of C. elegans appears four times, which is significantly higher than that of the human (which appears only once) in PMID: 18627611. The case led some human genes to be wrongly assigned to C. elegans. The remainder of the errors are relatively small and are caused by various factors, e.g., one gene mention for multiple species.

Also, as shown in Table 2, 767 (1245 – 478 = 767) documents in GNormPlus and NLM-Gene corpora were not used to benchmark, as the deep learning approach is not applicable. These articles can be grouped into three types: (1) No species can be found in the article, and thus, the straightforward method is to assign all of the genes to humans; (2) only one species is found in the article, and, thus, all of the genes can be assigned to the species without using the deep learning method; and (3) the corresponding species of the genes are not in abstract, and, thus, we can assign the remaining species in the article only to the genes, but none is correct. When we apply simple rules, we attain a comparable accuracy of 85.4% on these documents.

## 5  CONCLUSION

In this manuscript, we first proposed a species tagger with state-of-the-art performance and further presented a novel idea to address species assignment, which is the biggest challenge of the gene normalization task. The task of recognizing the corresponding species from various candidates for gene mentions is more relevant to information-retrieval or relation-extraction tasks, but we rephrased the problem into a sequence-labeling task, which is normally applied to a named-entity recognition task. The new method raised the performance accuracy of the species assignment (from 65.8% to 81.3%) within an acceptable process speed for large-scale data processing. Based on these promising results, we believe that the sequence labeling framework of species assignment can work with other relevant topics as well (e.g., the corresponding genes/species of variants and the corresponding variants of the phenotypes). Nevertheless, the tool is currently being developed and evaluated only on abstracts. Because more detailed information is recorded in the full text, in the future we would like to further improve our methods to be able to handle full-length articles with multiple passages as well as the highly unstructured parts (e.g., tables) of the text.

**ACKNOWLEDGMENTS**

This research was supported by the Intramural Research Program of the National Library of Medicine (NLM), National Institutes of Health.

**REFERENCES**

[1] Sylvain Poux, Cecilia N Arighi, Michele Magrane, Alex Bateman, Chih-Hsuan Wei, Zhiyong Lu, Emmanuel Boutet, Hema Bye-A-Jee, Maria Livia Famiglietti, and Bernd Roechert. 2017. On expert curation and scalability: UniProtKB/Swiss-Prot as a case study. *Bioinformatics* 33, 21 (2017), 3454-3460.
[2] Cathy H Wu, Cecilia N Arighi, Kevin B Cohen, Lynette Hirschman, Martin Krallinger, Zhiyong Lu, Carolyn Mattingly, Alfonso Valencia, Thomas C Wiegers, and W John Wilbur. 2012. BioCreative-2012 virtual issue. *Database* 2012, (2012).
[3] Qingyu Chen, Alexis Allot, and Zhiyong Lu. 2021. LitCovid: an open database of COVID-19 literature. *Nucleic acids research* 49, D1 (2021), D1534-D1540.
[4] Alexis Allot, Yifan Peng, Chih-Hsuan Wei, Kyubum Lee, Lon Phan, and Zhiyong Lu. 2018. LitVar: a semantic search engine for linking genomic variant data in PubMed and PMC. *Nucleic Acids Research* 46, W1 (2018), W530–W536.

[5] Kyubum Lee, Maria Livia Famiglietti, Aoife McMahon, Chih-Hsuan Wei, Jacqueline Ann Langdon MacArthur, Sylvain Poux, Lionel Breuza, Alan Bridge, Fiona Cunningham, Ioannis Xenarios, and Zhiyong Lu 2018. Scaling up data curation using deep learning: An application to literature triage in genomic variation resources. *PLoS computational biology* 14, 8 (2018), e1006390.

[6] Chih-Hsuan Wei, Alexis Allot, Robert Leaman, and Zhiyong Lu. 2019. PubTator central: automated concept annotation for biomedical full text articles. *Nucleic acids research* 47, W1 (2019), W587-W593.

[7] Chih-Hsuan Wei, Hung-Yu Kao, and Zhiyong Lu. 2015. GNormPlus: an integrative approach for tagging genes, gene families, and protein domains. *BioMed research international* 2015, (2015).

[8] Zhiyong Lu, Hung-Yu Kao, Chih-Hsuan Wei, Minlie Huang, Jingchen Liu, Cheng-Ju Kuo, Chun-Nan Hsu, Richard Tzong-Han Tsai, Hong-Jie Dai, and Naoaki Okazaki. 2011. The gene normalization task in BioCreative III. *BMC bioinformatics* 12, 8 (2011), S2.

[9] Jörg Hakenberg, Martin Gerner, Maximilian Haeussler, Illés Solt, Conrad Plake, Michael Schroeder, Graciela Gonzalez, Goran Nenadic, and Casey M Bergman. 2011. The GNAT library for local and remote gene mention normalization. *Bioinformatics* 27, 19 (2011), 2769-2771.

[10] Martin Gerner, Goran Nenadic, and Casey M Bergman. 2010. LINNAEUS: a species name identification system for biomedical literature. *BMC bioinformatics* 11, 1 (2010), 1-17.

[11] Nona Naderi, Thomas Kappler, Christopher JO Baker, and René Witte. 2011. OrganismTagger: detection, normalization and grounding of organism entities in biomedical documents. *Bioinformatics* 27, 19 (2011), 2721-2729.

[12] Evangelos Pafilis, Sune P Frankild, Lucia Fanini, Sarah Faulwetter, Christina Pavloudi, Aikaterini Vasileiadou, Christos Arvanitidis, and Lars Juhl Jensen. 2013. The SPECIES and ORGANISMS resources for fast and accurate identification of taxonomic names in text. *PLoS One* 8, 6 (2013), e65390.

[13] Karin Verspoor, Christophe Roeder, Helen L Johnson, Kevin Bretonnel Cohen, William A Baumgartner Jr, and Lawrence E Hunter. 2010. Exploring species-based strategies for gene normalization. *IEEE/ACM Transactions on Computational Biology Bioinformatics and Biology Insights* 7, 3 (2010), 462-471.

[14] Minlie Huang, Jingchen Liu, and Xiaoyan %J Bioinformatics Zhu. 2011. GeneTUKit: a software for document-level gene normalization. 27, 7 (2011), 1032-1033.

[15] Chih-Hsuan Wei, Hung-Yu Kao, and Zhiyong Lu. 2012. SR4GN: a species recognition software tool for gene normalization. In *the 8th Annual NIH Graduate Student Research Symposium*.

[16] Yu Gu, Robert Tinn, Hao Cheng, Michael Lucas, Naoto Usuyama, Xiaodong Liu, Tristan Naumann, Jianfeng Gao, and Hoifung Poon. 2021. Domain-specific language model pretraining for biomedical natural language processing. *ACM Transactions on Computing for Healthcare* 3, 1 (2021), 1-23.

[17] Li Fang, and Kai Wang. 2021. Team Bioformer at BioCreative VII LitCovid Track: Multic-label topic classification for COVID-19 literature with a compact BERT model. In *BioCreative VII workshop*.

[18] Leon Weber, Mario Sänger, Jannes Münchmeyer, Maryam Habibi, Ulf Leser, and Alan Akbik. 2021. HunFlair: an easy-to-use tool for state-of-the-art biomedical named entity recognition. *Bioinformatics* 37, 17 (2021), 2792-2794.

[19] Carl von Linné. 1735. *Systema naturae; sive, Regna tria naturae: systematice proposita per classes, ordines, genera & species*: Haak.

[20] Zhiheng Li, Zhihao Yang, Yang Xiang, Ling Luo, Yuanyuan Sun, and Hongfei Lin. 2020. Exploiting sequence labeling framework to extract document-level relations from biomedical texts. *BMC bioinformatics* 21, 1 (2020), 1-14.

[21] Ling Luo, Zhihao Yang, Mingyu Cao, Lei Wang, Yin Zhang, and Hongfei Lin. 2020. A neural network-based joint learning approach for biomedical entity and relation extraction from biomedical literature. *Journal of biomedical informatics* 103, (2020), 103384.

[22] Ling Luo, Po-Ting Lai, Chih-Hsuan Wei, and Zhiyong Lu. 2021. Extracting Drug-Protein Interaction using an Ensemble of Biomedical Pre-trained Language Models through Sequence Labeling and Text Classification Techniques. In *Proceedings of the BioCreative VII challenge evaluation workshop.* 26-30.

[23] John M Giorgi, and Gary D Bader. 2018. Transfer learning for biomedical named entity recognition with neural networks. *Bioinformatics* 34, 23 (2018), 4087-4094.

[24] Jinhyuk Lee, Wonjin Yoon, Sungdong Kim, Donghyeon Kim, Sunkyu Kim, Chan Ho So, and Jaewoo Kang. 2020. BioBERT: a pre-trained biomedical language representation model for biomedical text mining. *Bioinformatics* 36, 4 (2020), 1234-1240.

[25] Leon Weber, Jannes Münchmeyer, Tim Rocktäschel, Maryam Habibi, and Ulf Leser. 2020. HUNER: improving biomedical NER with pretraining. *Bioinformatics* 36, 1 (2020), 295-302.

[26] Rezarta Islamaj, Chih-Hsuan Wei, David Cissel, Nicholas Miliaras, Olga Printseva, Oleg Rodionov, Keiko Sekiya, Janice Ward, and Zhiyong Lu. 2021. NLM-Gene, a richly annotated gold standard dataset for gene entities that addresses ambiguity and multi-species gene recognition. *Journal of Biomedical Informatics* 118, (2021), 103779.

[27] Diederik P Kingma, and Jimmy Ba. 2015. Adam: A method for stochastic optimization. In *The International Conference on Learning Representations (ICLR).*